\documentclass[letterpaper]{article} 
\usepackage{aaai24}  
\usepackage{times}  
\usepackage{helvet}  
\usepackage{courier}  
\usepackage[hyphens]{url}  
\usepackage{graphicx} 
\usepackage{natbib}  
\usepackage{caption} 
\usepackage{booktabs}
\usepackage{adjustbox}
\usepackage[usenames,dvipsnames]{xcolor}
\usepackage{algorithm}
\usepackage{algorithmic}

\usepackage{newfloat}
\usepackage{listings}
\usepackage{subcaption}
\usepackage{rotating}
\usepackage{multirow}
\newcommand{\ie}{\textit{i}.\textit{e}., }
\newcommand{\eg}{\textit{e}.\textit{g}., }

\newcommand{\answerYes}[1]{\textcolor{blue}{#1}} 
 
\newcommand{\answerNA}[1]{\textcolor{gray}{#1}} 
 
\DeclareCaptionStyle{ruled}{labelfont=normalfont,labelsep=colon,strut=off} 
\floatstyle{ruled}
\newfloat{listing}{tb}{lst}{}
\floatname{listing}{Listing}
\title{Computational Assessment of Hyperpartisanship in News Titles}
\author{
    Hanjia Lyu\equalcontrib,
    Jinsheng Pan\equalcontrib,
    Zichen Wang\equalcontrib,
    Jiebo Luo}
\affiliations {
    University of Rochester\\
    \{hlyu5, jpan24, zwang189\}@ur.rochester.edu, jluo@cs.rochester.edu
}

\usepackage{bibentry}

\begin{document}

\maketitle

\begin{abstract}
The growing trend of partisanship in news reporting can have a negative impact on society. Assessing the level of partisanship in news headlines is particularly crucial, as they are easily accessible and frequently provide a condensed summary of the article's opinions or events. Therefore, they can significantly influence readers' decision to read the full article, making them a key factor in shaping public opinion. 
We first adopt a human-guided machine learning framework to develop a new dataset for \textit{hyperpartisan news title detection} with 2,200 manually labeled and 1.8 million machine-labeled titles that were posted from 2014 to the present by nine representative media organizations across three media bias groups - {\tt Left}, {\tt Central}, and {\tt Right} in an active learning manner. 
A fine-tuned transformer-based language model achieves an overall accuracy of 0.84 and an $F_1$ score of 0.78 on an external validation set. 
Next, we conduct a computational analysis to quantify the \textbf{extent and dynamics} of partisanship in news titles. While some aspects are as expected, our study reveals new or nuanced differences between the three media groups. 
We find that overall the {\tt Right} media tends to use proportionally more hyperpartisan titles. Roughly around the 2016 Presidential Election, the proportions of hyperpartisan titles increased across all media bias groups, with the {\tt Left} media exhibiting the most significant relative increase. 
We identify three major topics including \textit{foreign issues}, \textit{political systems}, and \textit{societal issues} that are suggestive of hyperpartisanship in news titles using logistic regression models and the Shapley values. 
Through an analysis of the topic distribution, we find that societal issues gradually gain more attention from all media groups. 
We further apply a lexicon-based language analysis tool to the titles of each topic and quantify the linguistic distance between any pairs of the three media groups, uncovering three distinct patterns. Codes and data are available at \url{https://github.com/VIStA-H/Hyperpartisan-News-Titles}.
\end{abstract}

\section{Introduction}

Exposure to hyperpartisan news online such as the coverage of growing polarization could lead individuals to perceive that both the political system and the public are highly polarized~\cite{fiorina2005culture}. In this study, we aim to understand the partisanship in news. To facilitate a quantitative understanding of \textbf{the extent of partisanship}, \citet{vincent2018crowdsourced} and \citet{kiesel-etal-2019-semeval} curated a \textit{hyperpartisan news} dataset. According to their definition, hyperpartisan articles ``mimic the form of regular news articles, but are one-sided in the sense that opposing views are either ignored or fiercely attacked''.  In the same vein, \citet{potthast-etal-2018-stylometric} considered hyperpartisan news as ``typically extremely one-sided, inflammatory, emotional, and often riddled with untruths.''

We believe that a more comprehensive understanding of hyperpartisanship can be achieved by considering not only \textbf{(1) the news that contains one-sided opinions} but also \textbf{(2) the news that describes conflicts and the underlying politically polarized climate} because both of them could lead to an increase in the public's perceived polarization~\cite{yang2016others, fiorina2005culture, levendusky2016does}. Additionally, the quantity of coverage itself can be considered as a particular form of bias~\cite{lin2011more}. Specifically, we seek to extend the definitions of hyperpartisan news to \textbf{include news that covers partisan conflicts and confrontations}. 

To this end, we first develop a dataset to measure the partisanship in \textbf{news titles}. We focus on titles because they are easily accessible to online users and often summarize the opinions or events of the full article, thereby influencing the decision to read the news content.
Furthermore, carefully crafted headlines achieve an optimal balance between capturing the reader's attention and conveying the key message with minimal cognitive effort, thus enabling readers to easily comprehend and synthesize the information presented~\cite{dor2003newspaper}. Titles that are readily available and circulated in like-minded online communities tend to become anchor points for individual opinions~\cite{yang2016others,zillmann1999exemplification}, especially when individuals do not directly experience the covered events~\cite{busselle2003media}.

\citet{yang2016others} conducted a questionnaire-based study on the relationship between media use and perceived political polarization. However, one of the limitations is the lack of examinations in real-world social platforms where media exposure is more complicated because of the combination of platforms' recommendation algorithms and sharing behaviors of friends and followers~\cite{bakshy2015exposure}. In other words, respondents may behave differently when they are finishing a questionnaire than when they are exposed to news on social platforms without knowing they are surveyed~\cite{krumpal2013determinants}. A study on the relationship between their opinions and hyperpartisanship in news titles can be conducted given a dataset that quantitatively measures hyperpartisanship in news titles. Therefore, we first develop a new dataset that complements previous hyperpartisan news datasets and then conduct an initial quantitative comparative analysis of more generalized hyperpartisan news titles. To summarize, our study consists of two parts:

\begin{itemize}
    \item \textbf{Hyperpartisan title detection} Based on the experimental results of our preliminary study on an existing hyperpartisan news dataset (details in \textit{Material and Method}), we find that \textit{class imbalance}, \textit{task-label misalignment}, and \textit{distribution shift} issues may hinder us from directly applying it to hyperpartisan title detection. Therefore, we adopt a human-guided machine learning framework to develop \textbf{a new hyperpartisan news titles dataset} with 2,200 manually labeled and 1.8 million machine-labeled titles that were posted by nine representative media organizations across three media bias groups - {\tt Left}, {\tt Central}, and {\tt Right}. The detection achieves an overall accuracy of 0.84 and an $F_1$ score of 0.78. We build it upon the ideas of a previously proposed hyperpartisan news dataset~\cite{kiesel-etal-2019-semeval,vincent2018crowdsourced} and further include a second type of news that could also affect perceived polarization. Another distinction between our dataset and the previous dataset is that we only focus on titles instead of full articles. 
    \item \textbf{Computational analysis of hyperpartisan titles} We conduct a computational analysis of 1.8 million news titles that were posted by nine representative media organizations from 2014 to 2022. For simplicity, we categorize and refer to them as {\tt Left}, {\tt Central}, and {\tt Right} media. However, additional attention should be paid when interpreting the results as they are based on nine instead of all media organizations. We find that overall {\tt Right} media tends to use proportionally more hyperpartisan titles, followed by {\tt Left} media and then {\tt Central} media. Moreover, roughly before the 2016 Presidential Election, a rise in the proportion of hyperpartisan titles is observed in all three media bias groups. In particular, the relative increase in the proportion of hyperpartisan titles of {\tt Left} is the greatest, while the relative changes of {\tt Right} and {\tt Central} are more similar and moderate. After Biden was elected the $46^{th}$ President, the proportions of hyperpartisan titles dropped and seemed to gradually return to the level before the 2016 Presidential Election. Moreover, three topics including \textit{foreign issues}, \textit{political systems}, and \textit{societal issues} are important in understanding the usage of hyperpartisan titles. A topic divergence including the focus on the choice of topics and linguistic differences is further observed among the media groups.
\end{itemize}

In the subsequent sections, we provide an overview of related work, discuss more details about our motivations, describe the development and labeling of our dataset in \textit{Material and Method}, present the results of our computational analysis, and discuss these results, limitations, and directions for future research. Finally, we explore broader perspectives and ethical considerations.

\section{Related Work}
News reports are not always free from bias. For instance, \citet{bourgeois2018selection} discovered selection biases in news coverage. Additionally, political partisanship bias is widespread in news media, with headlines, article sizes, and framings often manipulated strategically to align with particular ideologies~\cite{d2000media, budak2016fair, gentzkow2010drives}. This can lead to skewed reporting, making it crucial for readers to be mindful of this potential bias when consuming news.

Understanding partisanship in texts is an important research topic as it is closely related to the studies of political polarization~\cite{monroe2008fightin,gentzkow2010drives}. One of its subtasks - hyperpartisan news detection, has drawn great attention from the research community. \citet{potthast-etal-2018-stylometric} conducted an analysis of the writing style of 1,627 manually labeled hyperpartisan news articles and found that the writing style can help discriminate hyperpartisan news from non-hyperpartisan news. \citet{kiesel-etal-2019-semeval} raised a contest in hyperpartisan news detection. They developed a hyperpartisan news dataset with 1,273 manually labeled and 754,000 machine-labeled news articles. Forty-two teams completed the contest. Deep learning methods as well as the more traditional models that use hand-crafted features show promising performance. 

More recently, researchers have discovered the dominant performance of transformer-based pre-trained language models in a variety of natural language processing tasks~\cite{DBLP:conf/naacl/DevlinCLT19, chen2021fine, zhang2021monitoring}. \citet{zhang2021monitoring} created a fusion classifier that combines deep learning model scores, other psychological textual features, and user demographics to detect depression signals in social media posts. \citet{card2022computational} applied RoBERTa~\cite{liu2019roberta} to 200,000 congressional speeches and 5,000 presidential communications regarding immigration and discovered a significant shift towards more positive language in political discourse around immigration, compared to the past.

In this study, we intend to leverage a transformer-based language model to quantify the \textbf{extent and dynamics} of hyperpartisanship in \textbf{news titles} at a larger scale (\ie 1.8 million titles of nine representative news media from 2014 to 2022) compared to previous studies. Furthermore, while most studies focus on the comparison between left-leaning and right-leaning media, we expand our analysis to include the comparison with central media.

\section{Material and Method}
\subsection{Hyperpartisan Title Detection}
\subsubsection{Task Definition}
Inspired by \citet{kiesel-etal-2019-semeval}, we define hyperpartisan title detection as ``\textit{Given the title of an online news article, decide whether the title is hyperpartisan or not}''. A news title is considered hyperpartisan if it either (1) expresses a one-sided opinion (\eg denouncement, criticism) against a policy, a political party, or a politician in a biased and aggressive tone, or (2) describes a confrontation or a conflict between opposing parties indicating the underlying political climate is polarized. 

\begin{table}[t]
    \centering
    \adjustbox{max width=\linewidth}{
\begin{tabular}{c l cccc}
\toprule[1.1pt]
 & \multicolumn{1}{c}{Case}                & Accuracy & Precision & Recall & $F_1$   \\
\midrule
\textbf{1} & Original & 0.51     & 0.36      & 0.28   & 0.51 \\
\textbf{2} & Downsampling     & 0.52     & 0.48      & 0.44   & 0.53 \\
\textbf{3} & Testing year = 2016       & 0.47     & 0.38      & 0.92   & 0.61 \\
\textbf{4} & Testing year = 2017      & 0.48     & 0.87      & 0.47   & 0.61 \\
\textbf{5} & Testing year = 2018      & 0.66     & 0.67      & 0.59   & 0.63 \\
\bottomrule[1.1pt]
\end{tabular}}
    \caption{Hyperpartisan title detection on SemEval.}
    \label{tab:performance_semeval}
\end{table}

\subsubsection{Motivating Analysis}
The most closely related task to our goal is \textit{hyperpartisan news detection} as we have discussed in \textit{Introduction}. The hyperpartisan news dataset curated by \citet{vincent2018crowdsourced} and \citet{kiesel-etal-2019-semeval} contains 1,273 manually labeled news articles that were published by both active hyperpartisan and mainstream media outlets. The articles were labeled 1 if they contained a fair amount of hyperpartisan content, or 0 if not. For simplicity, we refer to this dataset as \textbf{SemEval}. One primary reason we cannot directly apply it to our task is that it only considers one type of hyperpartisan news which mainly conveys one-sided opinions. However, we aim to include an extra type of hyperpartisan news that describes an underlying politically polarized climate. Moreover, we perform a preliminary analysis to show that there are \textbf{three more major issues} that prevent us from directly using it for hyperpartisan title detection. 

\begin{itemize}
    \item \textbf{Class imbalance} Following the same train-test split of SemEval, we fine-tune a pre-trained BERT-base model~\cite{DBLP:conf/naacl/DevlinCLT19} to predict whether a news title is hyperpartisan based \textit{solely} on the title itself. However, it achieves a poor performance (Row 1 of Table~\ref{tab:performance_semeval}). We hypothesize that one potential cause is the class imbalance issue as the positive samples only account for 36.90\% in the training set. In Row 2, we downsample the data of the majority class in the training set and conduct the same experiment again. We observe that the performance improves when the imbalance issue is alleviated.

    \item \textbf{Task-label misalignment} Even after we apply downsampling to SemEval, the improvement is still limited. One reason we hypothesize is that the task and label are \textbf{not} aligned. Our task is to predict whether a news title is hyperpartisan, however, the labels in SemEval are rated based on the full article and its metadata (\eg publisher). 
    \item \textbf{Distribution shift} A majority of the articles of SemEval were posted from 2016 to 2018. To verify if  hyperpartisan title detection is robust across different years, we further conduct three experiments. For each experiment, we use the articles from one year as the testing set, and all the other articles as the training set. The same fine-tuning process is performed. Rows 3, 4, and 5 of Table~\ref{tab:performance_semeval} show the performance of using the articles posted in 2016, 2017, and 2018, respectively, as the testing set. The results suggest a poor and inconsistent performance across different years. It may be because of a distribution shift in the textual patterns such as tones, context, and so on (which is later confirmed by our analysis results). 
\end{itemize}

The experimental results provide insights into the construction of our dataset for hyperpartisan news title detection. The dataset should (1) be class-balanced in order to support robust learning for language models~\cite{japkowicz2000learning,chawla2004special}; (2) have labels that are rated solely based on titles; and (3) contain titles of different years so that the prediction performance is consistent.  Most importantly, the dataset should also consider both two types of hyperpartisan news. Next, we develop a dataset that satisfies these conditions.

\subsubsection{Data collection}
We collect news from nine representative media outlets of different media biases - {\tt Left}, {\tt Central}, and {\tt Right}.  The media bias is judged and assigned by allsides.com and mediabiasfactcheck.com. The New York Times, CNN, NBC, and Bloomberg are included in {\tt Left} media. Wall Street Journal and Christian Science Monitor are included in {\tt Central} media. The Federalist, Reason, and Washington Times are included in {\tt Right} media. Fox is not included because access to its history news titles is not publicly available. Two methods are employed to collect news: (1) using the official web API provided by the news media, and (2) crawling the archive/sitemap page that is publicly available on the media's official websites. News from The New York Times and Bloomberg is collected using the first approach. News from the remaining media outlets is collected using the second approach. As aforementioned, we focus on the titles of the articles. In total, we have collected over 1.8 million titles from January 2014 to September 2022.

\subsubsection{Human-guided machine learning}
After a preliminary exploration of the collected titles, we observe a class imbalance issue, with only a small number of titles being hyperpartisan. Consequently, we will have to label more data to have enough hyperpartisan samples. To label titles more efficiently, we develop our dataset in an \textbf{active learning} manner, where a machine learning model and human annotators are guided by each other to iteratively seek samples of the minority class and improve the performance of the machine learning model. We adopt this strategy because it has shown effectiveness in developing datasets and handling class imbalance in multiple studies~\cite{sadilek2013nemesis,lyu2022social,zhou2022ban}.

\begin{figure}[t]
    \centering
    \includegraphics[width=\linewidth]{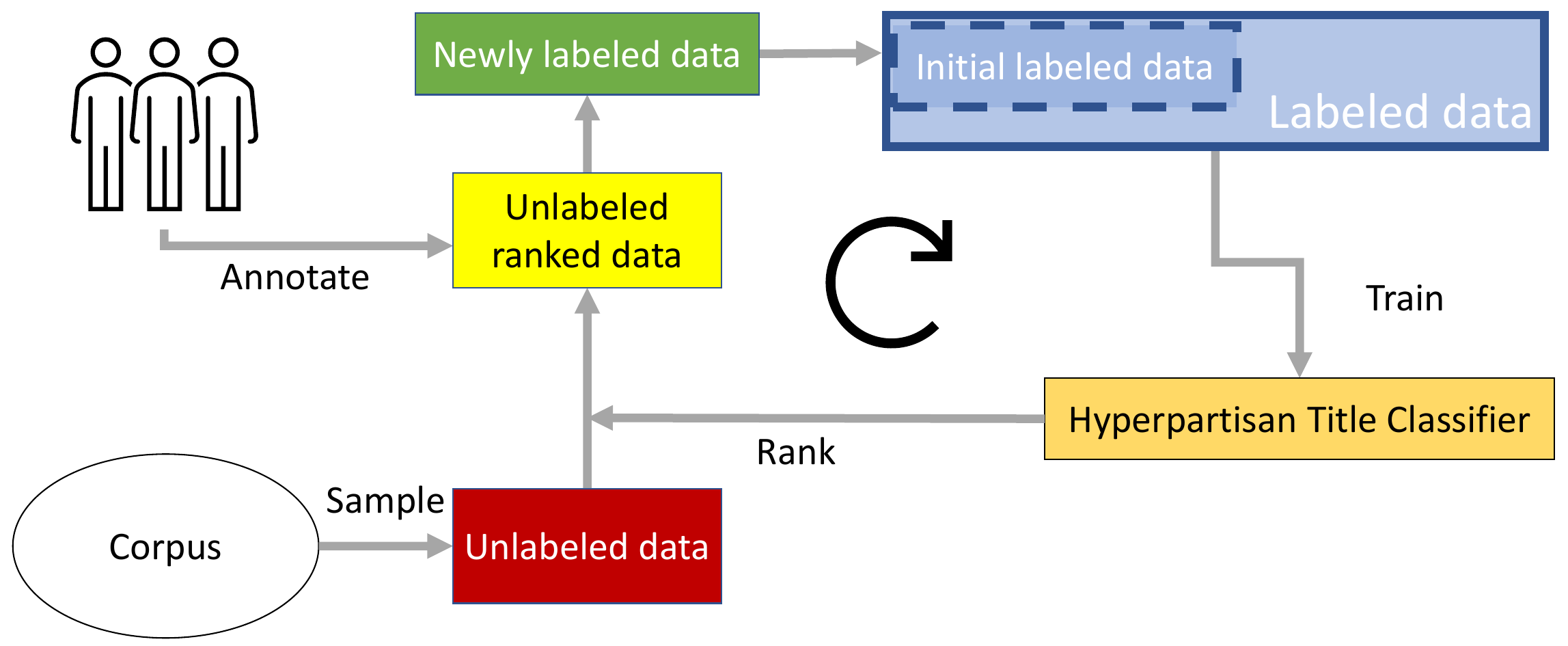}
    \caption{The human-guided machine learning framework.}
    \label{fig:active_learning}
\end{figure}

\begin{table*}[t]
\centering
\setlength{\tabcolsep}{5pt}
\small
\begin{tabular}{l|l|ccccccccc}
\toprule[1.1pt]
                                  &                                    & 2014 & 2015 & 2016 & 2017 & 2018 & 2019 & 2020 & 2021 & 2022 \\ \midrule
\multirow{2}{*}{\begin{tabular}[c]{@{}l@{}}Manually \\ labeled\end{tabular}} & \# hyperpartisan titles     & 71   & 84     & 87     & 119     & 124     & 114     & 116     & 105     & 78     \\
                                  & \# non-hyperpartisan titles & 149     & 173     & 171     & 143     & 130     & 132     & 140     & 127     & 137     \\ \midrule
\multirow{2}{*}{\begin{tabular}[c]{@{}l@{}}Model \\ labeled\end{tabular}}    & \# hyperpartisan titles     &      24,258 & 24,706 & 34,626 & 34,181 & 32,259 & 33,326 & 34,766 & 28,075 & 22,495     \\
                                  & \# non-hyperpartisan titles &      229,029 & 207,176 & 193,419 & 162,657 & 155,651 & 145,964 & 155,173 & 160,403 & 144,460    \\
                                  \bottomrule[1.1pt]
\end{tabular}
 \caption{Summary statistics of the dataset. The total number of manually labeled titles is 2,200. The total number of titles that are labeled by the model is 1.8 million.}
    \label{tab:data_dist}
\end{table*}

\begin{table}[t]
\centering
\begin{tabular}{ccccc}
\toprule[1.1pt]
Iteration   & Accuracy & Precision & Recall & $F_1$ \\
\midrule
\textbf{1}          & 0.76        & 0.79          & 0.52       & 0.63   \\
\textbf{2}           & 0.78         & 0.77          & 0.62      & 0.67   \\
\textbf{3}           & 0.82         & 0.78          & 0.75       & 0.76   \\
\textbf{4}           & \textbf{0.84}         & \textbf{0.81}          & \textbf{0.76}       & \textbf{0.78}   \\
\bottomrule[1.1pt]
\end{tabular}
\caption{Performance of hyperpartisan title detection of each iteration of our model.}
\label{tab:performance_ours}
\end{table}

Figure~\ref{fig:active_learning} shows a diagram of the human-guided machine learning framework we adopt from \citet{lyu2022social}. In general, the learning process is composed of multiple iterations. In iteration $i$, annotators first manually label a small batch of data $D_{ranked\_unlabled}^{i}$. Next, we add this new batch of labeled data $D_{labled}^{i}$ into the labeled corpus $C_{train}$ and retrain the machine learning model $H$. The trained model is used to look for the samples of the minority class - hyperpartisan titles from corpus $C_{unlabeled}$. The samples that are rated as most likely hyperpartisan by the model make up $D_{ranked\_unlabled}^{i+1}$ which will be manually labeled by the annotators in the next iteration. 

More specifically, $C_{unlabeled}$ is comprised of 1.8 million titles we collected. For each iteration, three annotators independently read a new batch of 500 titles and rate them as hyperpartisan or non-hyperpartisan. A news title is labeled hyperpartisan if (1) expresses a one-sided opinion (\eg denouncement, criticism) against a policy, a political party, or a politician in a biased and aggressive tone, or (2) describes a confrontation or a conflict between opposing parties indicating the underlying political climate is polarized. Otherwise, it is labeled non-hyperpartisan. After the annotation of the first iteration, the three annotators read each other's labeling results and discuss the disagreements which mainly come from the less familiarity with the language and politics of different periods. The final label is assigned with the consensus votes from three annotators. We choose the pre-trained BERT-base as our machine learning model $H$. 

In the first iteration, we fine-tune $H$ using SemEval and apply it to 2,000 titles we randomly sample from $C_{unlabled}$. These 2,000 titles are ranked based on the likelihood of being hyperpartisan predicted by $H$. The new batch is composed of 90\% ($n=450$) most likely hyperpartisan titles and 10\% ($n=50$) randomly sampled titles. The 10\% random samples are included for diversity. The specified proportions of the most likely hyperpartisan and random samples allow us actively look for the samples of the minority class which alleviates the \textbf{class imbalance} issue. Moreover, to address the potential \textbf{distribution shift} problem we discuss before, we add a time constraint. The 90\% ($n=450$) titles are composed of the 90\% most likely hyperpartisan titles of each year between 2014 to 2022. As a result, the number of potential hyperpartisan titles has a relatively uniform distribution across different years. In the following iterations, all settings are the same, except that we no longer need SemEval to fine-tune our model. Our model $H$ is fine-tuned iteratively with the aggregate labeled data from our own collected titles. 

We conduct the labeling and model training for four iterations. Inter-annotator agreement measured by Cohen's Kappa is 0.47, suggesting a moderate to substantial agreement~\cite{landis1977measurement}. More details on the annotation process can be found in the \textit{Appendix}.  We have also labeled an external validation set with 200 titles, with an equal split of 50\% hyperpartisan and 50\% non-hyperpartisan titles. Note that these 200 titles are from the same corpus of 1.8 million news titles but are labeled at the beginning of the labeling session. They are not used during the active learning process. The same annotation schema is applied. Table~\ref{tab:data_dist} shows the summary statistics of the 2,200 manually labeled titles. The performance of $H$ is evaluated on the external validation set at each iteration (Table~\ref{tab:performance_ours}). We find that the performance on hyperpartisan title detection is improved iteration by iteration. After four iterations, the model achieves an overall accuracy of 0.84 and an $F_1$ score of 0.78.

Moreover, we compare the performance of other representative language models with the BERT-base model (Table~\ref{tab:performance_model}). In general, the BERT-base model outperforms other language models. TF-IDF has a slightly higher recall score and a much lower precision score, suggesting poor performance. The superiority of the BERT-base model over bag-of-words, TF-IDF, and word2vec is \textit{consistent} with its performance in other classification tasks~\cite{DBLP:conf/naacl/DevlinCLT19, zhang2021monitoring, chen2021fine}. The preprocessing procedures, implementation details, and hyperparameter settings are discussed in the \textit{Appendix}.

It is noteworthy that the 1.8 million titles we collect are from nine media outlets. It is possible that the trained classifier may not generalize to other outlets. To verify the generalizability of our model, we conduct nine more experiments. For each experiment, the titles of one of the nine media outlets are completely removed from the training set and the testing set is only composed of the titles of this particular media outlet. 
The results, shown in Table~\ref{tab:bert_robustness}, demonstrate that the performance of the BERT-based model is \textit{consistent} and can generalize well to unseen outlets.

\begin{table}[t]
\centering
\begin{tabular}{ccccc}
\toprule[1.1pt]
Model   & Accuracy & Precision & Recall & $F_1$ \\
\midrule
Bag-of-words &  0.70       &  0.30         & 0.77       & 0.43   \\
TF-IDF & 0.73 & 0.39 & \textbf{0.79} & 0.52\\
word2vec    & 0.78         &  0.60         &  0.77      & 0.67  \\
BERT-base & \textbf{0.84}& \textbf{0.81}& 0.76& \textbf{0.78}\\
\bottomrule[1.1pt]
\end{tabular}
\caption{Performance of hyperpartisan title detection of different language models.}
\label{tab:performance_model}
\end{table}

\begin{table}[t]
\setlength{\tabcolsep}{3.8pt}
\small
    \centering
\begin{tabular}{ccccc}
\toprule[1.1pt]
Target Media                      & Accuracy & Precision & Recall & $F_1$   \\
\midrule
Bloomberg      & 0.83     & 0.79      & 0.78   & 0.78 \\
CNN & 0.84    & 0.80      & 0.77   & 0.78 \\
NBC     & 0.85     & 0.81     & 0.81   & 0.81 \\
The New York Times     &   0.84   & 0.77      & 0.86   & 0.81 \\
Christian Science Monitor       & 0.84     & 0.82      & 0.77   & 0.84 \\
Wall Street Journal      & 0.83     & 0.82      & 0.72   & 0.77 \\
Reason   & 0.84     & 0.85      & 0.73   & 0.78 \\
The Federalist       & 0.84     & 0.81      & 0.78   & 0.79 \\
Washington Time       & 0.85     & 0.82      & 0.81   & 0.81 \\
\bottomrule[1.1pt]
\end{tabular}
    \caption{Generalizability verification for our model.}
    \label{tab:bert_robustness}
\end{table}

\begin{figure*}[t]
\centering
\begin{subfigure}{0.45\textwidth}
    \includegraphics[width=\textwidth]{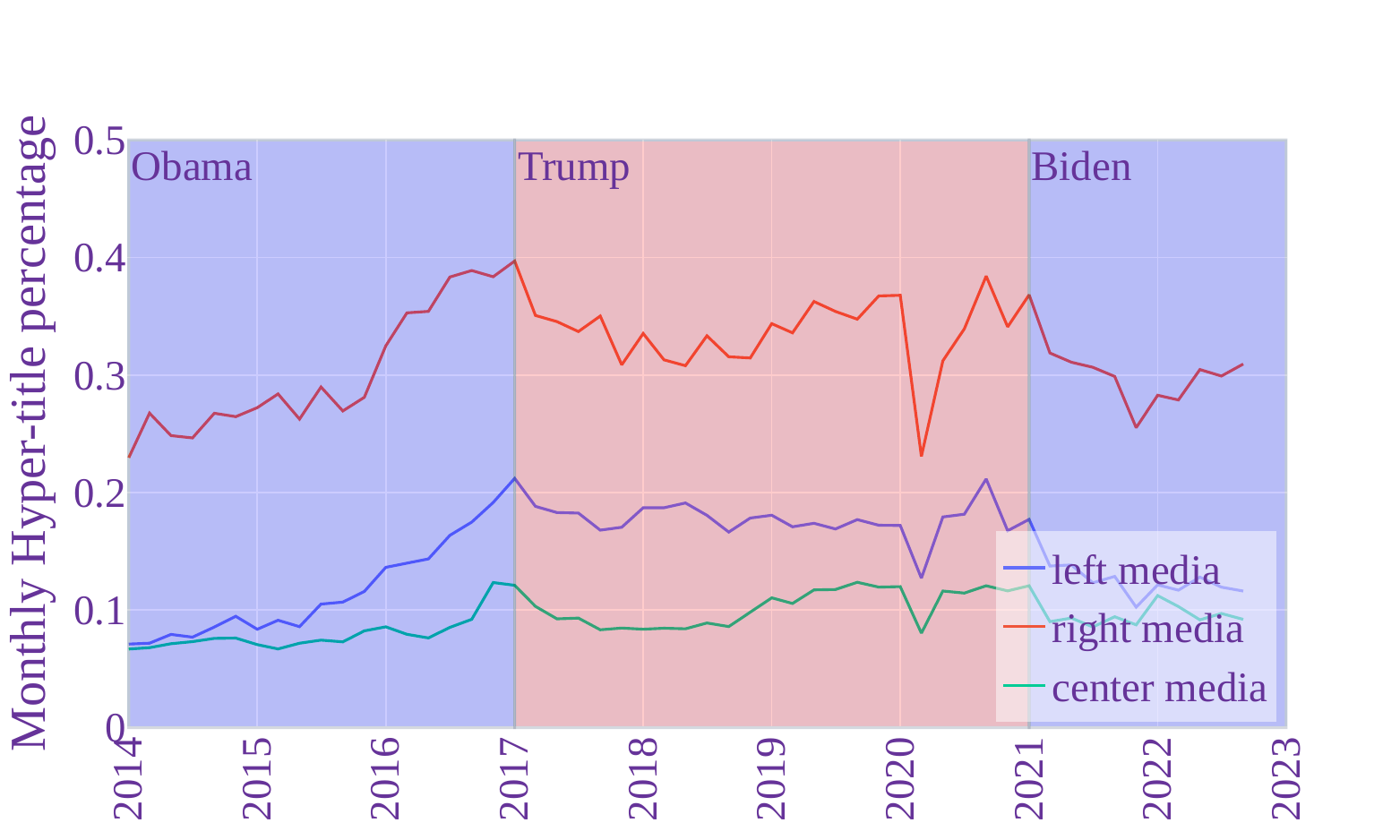}
    \caption{Percentage of hyperpartisan titles.}
    \label{fig:first_overall}
\end{subfigure}
\hfill
\begin{subfigure}{0.45\textwidth}
    \includegraphics[width=\textwidth]{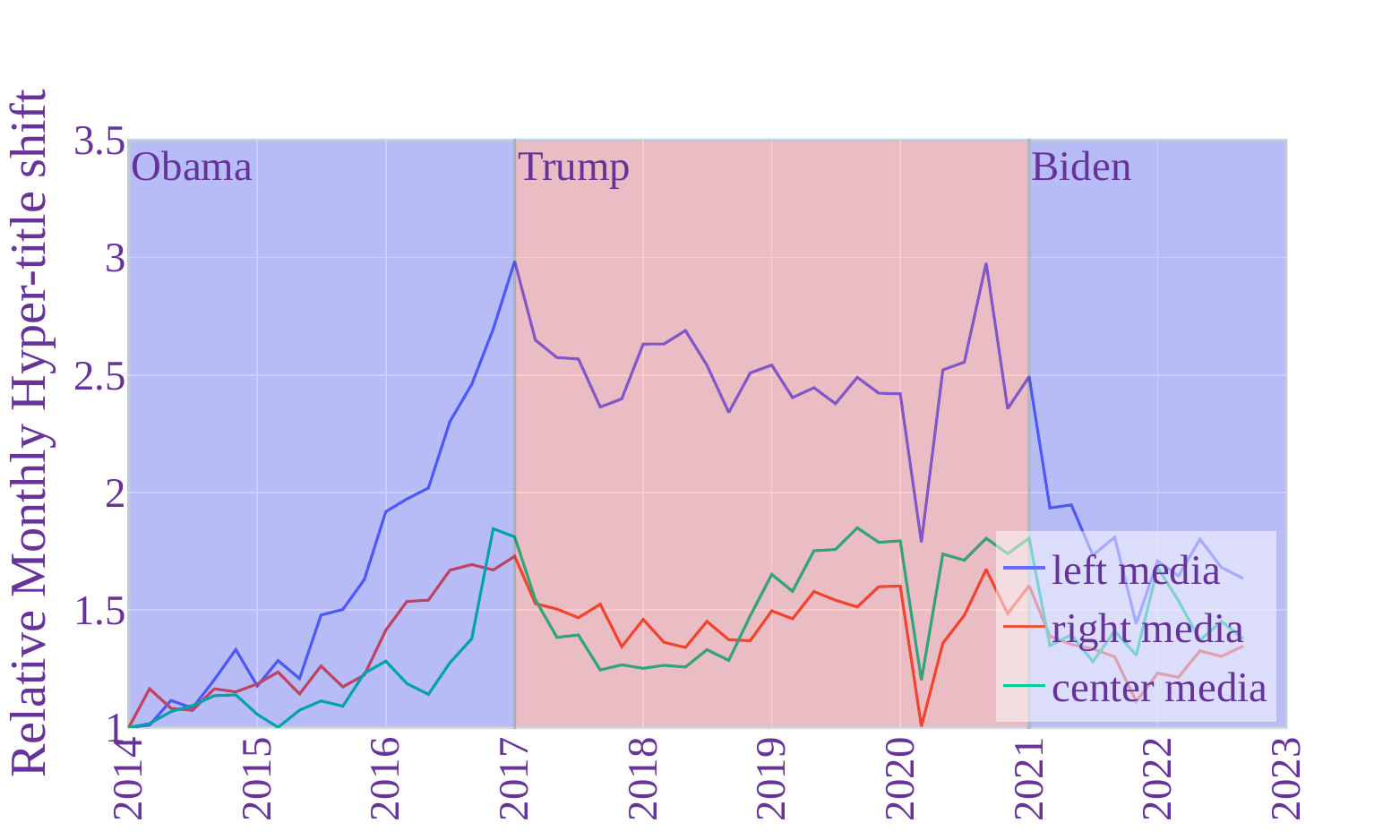}
    \caption{Relative change in percentage.}
    \label{fig:second_overall}
\end{subfigure}
\hfill
\begin{subfigure}{0.45\textwidth}
    \includegraphics[width=\textwidth]{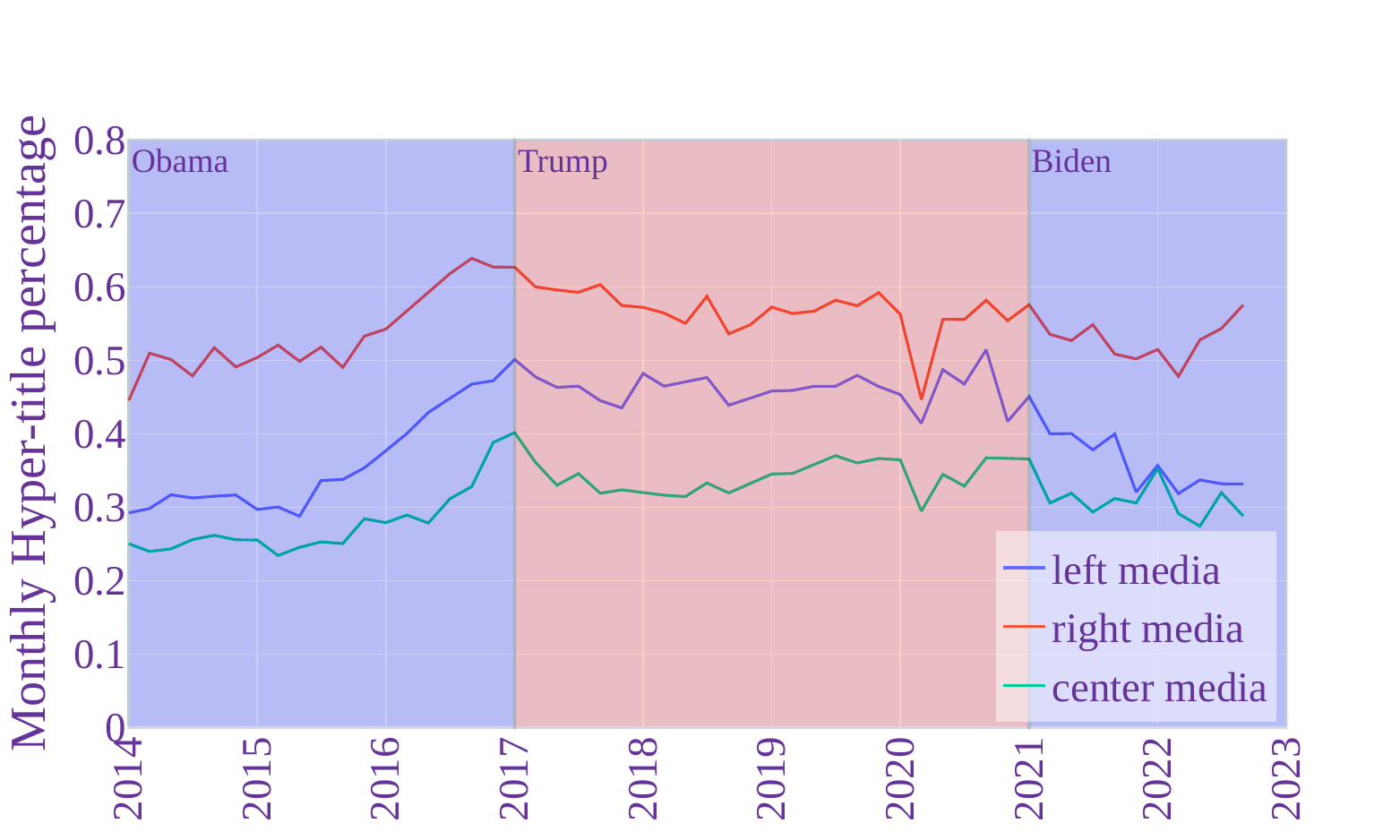}
    \caption{Percentage of hyperpartisan titles in the politics subset.}
    \label{fig:first_overall_political}
\end{subfigure}
\hfill
\begin{subfigure}{0.45\textwidth}
    \includegraphics[width=\textwidth]{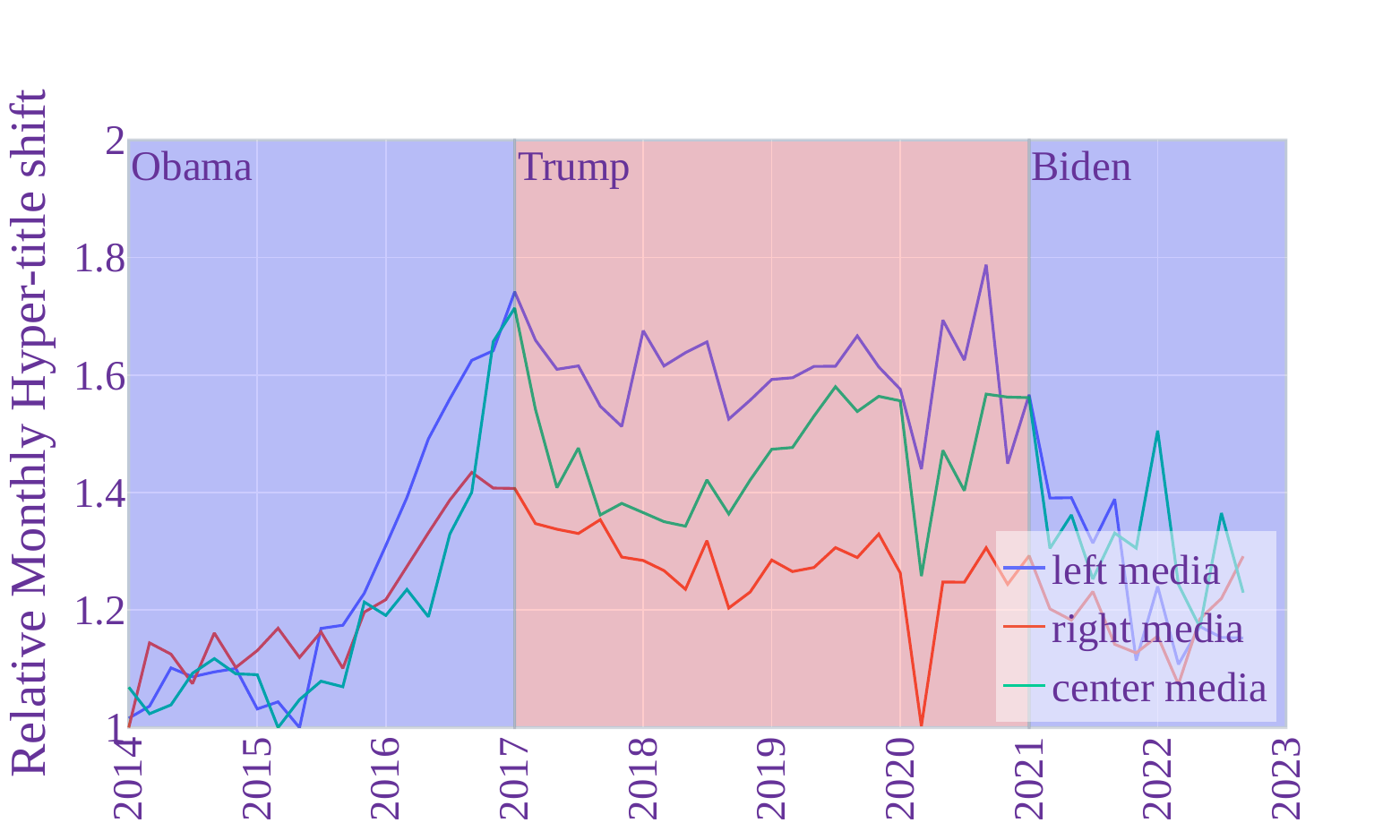}
    \caption{Relative change in percentage of the politics subset.}
    \label{fig:second_overall_political}
\end{subfigure}
\caption{Usage of hyperpartisan titles of Left, Central, and Right media from 2014 to 2022.}
\label{fig:overall}
\end{figure*}

\section{Results}
\subsection{Hyperpartisanship in News Titles}
The model is fine-tuned using our labeled data and applied to 1.8 million titles that were posted by {\tt Left}, {\tt Central}, and {\tt Right} media between 2014 and 2022. Table~\ref{tab:data_dist} shows the summary statistics of the titles that are labeled by our model. Overall, 14.7\% titles are classified as hyperpartisan. Figure~\ref{fig:first_overall} shows the percentage of hyperpartisan titles. To calculate the relative change in percentage, the percentage at each time stamp is divided by the initial percentage in 2014. The result is shown in Figure~\ref{fig:second_overall}. The \textit{Appendix} includes the results of an alternative model and a validity verification.

First, we find that overall {\tt Right} media tends to use proportionally more hyperpartisan titles, followed by {\tt Left} media and then {\tt Central} media (Figure~\ref{fig:first_overall}). Of all the titles between 2014 to 2022, 14.20\%, 8.91\%, and 31.65\% are hyperpartisan in {\tt Left}, {\tt Central}, and {\tt Right}, respectively. 

Second, we discover a rise in the proportion of hyperpartisan titles in all three media bias groups (Figure~\ref{fig:second_overall}). Specifically, the relative increase in the proportion of hyperpartisan titles of {\tt Left} media is the greatest, while the relative changes of {\tt Right} and {\tt Central} media are more similar and moderate. Compared to 2014, the average proportion of hyperpartisan titles that {\tt Left} used in 2016 increased by 101.77\% (22.19\% for {\tt Central} and 43.53\% for {\tt Right}). The proportions of hyperpartisan titles remained high during the Trump Administration (2017-2020).

Third, after Biden was elected the $46^{th}$ President, the proportions of hyperpartisan titles dropped and seemed to gradually return to the level before the 2016 Presidential Election. For instance, the relative change in hyperpartisan title proportion of {\tt Left} media dropped by 39.74\% by comparing November 2020 and November 2021.

Note that the increased proportion of hyperpartisan titles may be due to the increased coverage of political events such as presidential elections. To better understand the reason for the rise in the usage of hyperpartisan titles, we create a {\tt politics} subset which \textbf{only} contains titles regarding politics from the 1.8 million corpora. A separate labeling-training process is conducted (more details in the \textit{Appendix}). Figures~\ref{fig:first_overall_political} and \ref{fig:second_overall_political} show the percentage of hyperpartisan titles in the {\tt politics} subset and relative change in percentage of the {\tt politics} subset, respectively. We observe that the overall pattern \textbf{remains the same}. However, a notable difference is that although the relative change in percentage of {\tt Left} media is still the highest, it is smaller in the {\tt politics} subset than that in the overall set, suggesting that a great portion of the increased hyperpartisan titles posted by the {\tt Left} media is the coverage of political events.

\subsection{Influential Terms in Predicting Hyperpartisanship}
Similar to \citet{card2022computational}, we train L1 logistic regression models to identify the most important words that are suggestive of hyperpartisan and non-hyperpartisan titles used by {\tt Left}, {\tt Central}, and {\tt Right} media. Based on the observations in Figure~\ref{fig:overall}, we categorize the titles into three periods - before the 2016 Presidential Election, during the Trump Administration, and after the 2020 Presidential Election. The input to the logistic regression models is the words in each title. The output label - whether or not the title is hyperpartisan, is assigned based on the inference of our BERT-base model. We only include the words that at least appear in 0.5\% of all collected titles. Stop words are removed using the {\tt nltk} package. Word counts are binarized. 

The importance of each word is measured using the Shapley values~\cite{lundberg2017unified}. Each model is trained and evaluated using 5-fold cross-validation (Table~\ref{tab:mvalidity_regression}). In each validation, the 50 most important features (\ie words) are kept. After five validations, the features that occur the most are reported in Table~\ref{tab:impact_word}. Similar to the BERT-base model, logistic regression may suffer from the generalizability issue. To verify this, we conduct a set of additional experiments where the input of one media outlet is completely removed from the training set and the testing set is only composed of the input of this outlet. Table~\ref{tab:mvalidity_regression} shows that the logistic regression models generalize well to unseen outlets.
It is noteworthy that the accuracy appears to be lower for right-leaning outlets. This could be attributed to the limited training data available for titles from these outlets.

\begin{table*}[t]
    \centering
\begin{tabular}{lccc}
\toprule[1.1pt]
                          & \begin{tabular}[c]{@{}c@{}}Pre-2016 election\\ (2014-2016)\end{tabular} & \begin{tabular}[c]{@{}c@{}}Trump Administration\\ (2017-2020)\end{tabular} & \begin{tabular}[c]{@{}c@{}}Post-2020 election\\ (2021-2022)\end{tabular} \\
                          \midrule
 {\tt Left}                      &  0.92 $\pm$ 0.00                                                               & 0.89 $\pm$ 0.00                                                                  &  0.90 $\pm$ 0.00                                                                        \\
-  Bloomberg                 &  0.94 $\pm$ 0.00                                                               &  0.94 $\pm$ 0.00                                                                  &  0.92 $\pm$ 0.00                                                                \\
-  CNN                       &  0.86 $\pm$ 0.00                                                               &  0.85 $\pm$ 0.00                                                                  &  0.88 $\pm$ 0.00                                                                \\
-  NBC                       &  0.90 $\pm$ 0.00                                                               &  0.86 $\pm$ 0.00                                                                  &  0.86 $\pm$ 0.00                                                                \\
-  The New York Times        &  0.94 $\pm$ 0.00                                                               &  0.91 $\pm$ 0.00                                                                  &  0.91 $\pm$ 0.00                                                                \\
\midrule
 {\tt Central}                   &  0.93 $\pm$ 0.00                                                               &  0.92 $\pm$ 0.00                                                                  &  0.92 $\pm$ 0.00                                        \\
-  Christian Science Monitor &  0.88 $\pm$ 0.00                                                              &  0.88 $\pm$ 0.00                                                               &  0.85 $\pm$ 0.01                                                               \\
-  Wall Street Journal       &  0.94 $\pm$ 0.00                                                               &  0.93 $\pm$ 0.00                                                                  &  0.92 $\pm$ 0.00                                                                \\
\midrule
 {\tt Right}                     &                  0.79 $\pm$ 0.00                                                        &  0.79 $\pm$ 0.00                                                                          &       0.79 $\pm$ 0.00                                                                     \\
-  Reason                   &  0.79 $\pm$ 0.00                                                               &  0.81 $\pm$ 0.00                                                                  &  0.84 $\pm$ 0.00                                                                \\
-  The Federalist            &  0.77 $\pm$ 0.01                                                               &  0.75 $\pm$ 0.01                                                                &  0.70 $\pm$ 0.01                                                                \\
-  Washington Time           &  0.79 $\pm$ 0.01                                                               &  0.79 $\pm$ 0.00                                                              &  0.78 $\pm$ 0.00                                                  \\
\bottomrule[1.1pt]             
\end{tabular}
    \caption{Classification accuracy of the logistic regression models under 5-fold cross-validation. Each row indicated by the media name represents the classification accuracy of the generalizability verification. For instance, the row of Bloomberg reports the classification accuracy of the logistic regression model that is trained without the input from Bloomberg but is tested with the input from Bloomberg under 5-fold cross-validation.}
    \label{tab:mvalidity_regression}
\end{table*}

Our first observation is that across three periods, three groups of words representing foreign countries, domestic political systems, and societal issues are influential in predicting hyperpartisan and/or non-hyperpartisan titles. 

Since the pre-2016 election period only covers the news posted between 2014 to 2016, it is expected the words that are suggestive of either hyperpartisanship or non-hyperpartisanship are election-related (\eg ``campaign'' and ``debate''), although there is a subtle difference where in spite of the similarity in terms of the election, ``campaign'' and ``debate'' are more important in hyperpartisanship prediction while ``vote'' and ``election'' are more important in the non-hyperpartisan prediction of {\tt Left} media. Another noteworthy finding is that ``Biden'' is important in predicting hyperpartisan titles of {\tt Right} media while ``Trump'' is influential in predicting non-hyperpartisan titles.

During the Trump Administration, words indicating societal issues were more important in predicting hyperpartisan titles of {\tt Left} media. Interestingly, although COVID-19 was announced as a pandemic in 2020 (late in the Trump Administration), the related words (\ie ``coronavirus'' and ``covid'') are still influential at the aggregate level. Moreover, coverage of foreign issues becomes important in predicting hyperpartisanship of {\tt Right} media.

After the 2020 Presidential Election, some election-related words are still considered influential. However, it is not surprising as there is still an argument from Trump and his allies about the election result.

\begin{table*}[t]
\centering
\begin{tabular}{lllll}
\toprule[1.1pt]
                         &                   & \multicolumn{1}{c}{\begin{tabular}[c]{@{}c@{}}Pre-2016 election\\ (2014-2016)\end{tabular}}                                                                                           & \multicolumn{1}{c}{\begin{tabular}[c]{@{}c@{}}Trump Administration\\ (2017-2020)\end{tabular}}                                                                & \multicolumn{1}{c}{\begin{tabular}[c]{@{}c@{}}Post-2020 election\\ (2021-2022)\end{tabular}}                                                                       \\ \midrule
\multirow{5}{*}{\rotatebox[origin=c]{90}{Left}}    & \small\rotatebox[origin=c]{90}{H}     & \begin{tabular}[c]{@{}l@{}}power, Democrat, party, policy, \\ time, lead, crisis, star, push, \\ campaign, kill, Obama, \\ lose, GOP, debate\end{tabular}                     & \begin{tabular}[c]{@{}l@{}}party, star, government, election, \\ test, crisis, rate, gun, sale, \\ power, school, coronavirus, law, \\ world, woman\end{tabular} & \begin{tabular}[c]{@{}l@{}}Black, talk, kill, game, Democrat, \\ protest, campaign, call, right, \\ former, debate, group, power, \\ Russian, today\end{tabular}  \\ \cline{2-5} 
                         & \small\rotatebox[origin=c]{90}{Non-H} & \begin{tabular}[c]{@{}l@{}}game, thing, return, protest, law, \\ Black, Russian, White, company, \\ vote, rate, market, election,  woman\end{tabular}                 & \begin{tabular}[c]{@{}l@{}}Clinton, lose, rate, next, push, \\ man, kill, climate, live, senate, \\ Obama, debate, Republican, \\ company, attack\end{tabular}   & \begin{tabular}[c]{@{}l@{}}American, Republican, shoot, \\ attack, love, rate, climate, state, \\ covid, bank, Ukraine, America, \\ bill, buy, man\end{tabular}        \\\midrule
\multirow{4}{*}{\rotatebox[origin=c]{90}{Central}} & \small\rotatebox[origin=c]{90}{H}     & \begin{tabular}[c]{@{}l@{}}White, Ukraine, crisis, protest, \\ Democrat, debate, party, Black, \\ review, attack, government, \\ Republican, rate, supreme, Iran\end{tabular} & \begin{tabular}[c]{@{}l@{}}rate, Clinton, covid, right, lose, \\ call, senate, China, review, \\ Trump, push, north, bill, life, \\ sell\end{tabular}            & \begin{tabular}[c]{@{}l@{}}Texas, Republican, new, right, \\ Ukraine, car, kill, big, attack, \\ Trump, coronavirus, push, \\ Biden, GOP\end{tabular}         \\ \cline{2-5} 
                         & \small\rotatebox[origin=c]{90}{Non-H} & \begin{tabular}[c]{@{}l@{}}death, car, meet, game, election, \\ CEO, school, news, Korea, sale, \\ job, tech, test, star, trade\end{tabular}                                  & \begin{tabular}[c]{@{}l@{}}America, protest, cut, car, school, \\ Black, price, kill, supreme, work, \\ White, set, meet, stop, buy\end{tabular}                 & \begin{tabular}[c]{@{}l@{}}former, work, plan, child, vote, \\ look, covid, record, rate, ban, \\ tech, stock, policy, president, health\end{tabular}             \\ \midrule
\multirow{5}{*}{\rotatebox[origin=c]{90}{Right}}   & \small\rotatebox[origin=c]{90}{H}     & \begin{tabular}[c]{@{}l@{}}right, fight, record, seek, leave, \\ lose, book, risk, party, time, \\ school, push, protest, gun, Biden\end{tabular}                                      & \begin{tabular}[c]{@{}l@{}}book, big, test, review, attack, \\ stop, policy, Russian, tax, new, \\ Texas, Ukraine, star, court, war\end{tabular}               & \begin{tabular}[c]{@{}l@{}}power, want, state, protest, \\ pandemic, make, Democrat, \\ government, Black, Texas, time,\\ Clinton, ban, judge, former\end{tabular}  \\ \cline{2-5} 
                         & \small\rotatebox[origin=c]{90}{Non-H} & \begin{tabular}[c]{@{}l@{}}car, company, Ukraine, day, Trump, \\ Russian, market, return, America, \\ Black, leader, new, home, \\ government, old\end{tabular}                        & \begin{tabular}[c]{@{}l@{}}election, Democrat, work, home, \\ supreme, business, Obama, old, \\ return, win, White, take, time\end{tabular}                    & \begin{tabular}[c]{@{}l@{}}Biden, find, school, vote, election, \\ star, company, bill, party, start, \\ fight, meet, trade, old, seek\end{tabular}                 \\ \bottomrule[1.1pt]
\end{tabular}
\caption{Most influential words for hyperpartisan and non-hyperpartisan titles, in three periods, when approximating the predicted hyperpartisanship from our BERT-base model with logistic regression models. H: Hyperpartisan, Non-H: Non-hyperpartisan. We find words representing foreign countries (\eg ``Russia'' and ``China''), domestic political systems (\eg ``Democrat'' and ``Republican''), and societal issues (\eg ``gun'' and ``school''). }
\label{tab:impact_word}
\end{table*}

\subsection{Topic Divergence among Media Groups}
\subsubsection{Topic distribution}
\begin{figure*}[t]
\centering
\begin{subfigure}{0.33\textwidth}
    \includegraphics[width=0.85\textwidth]{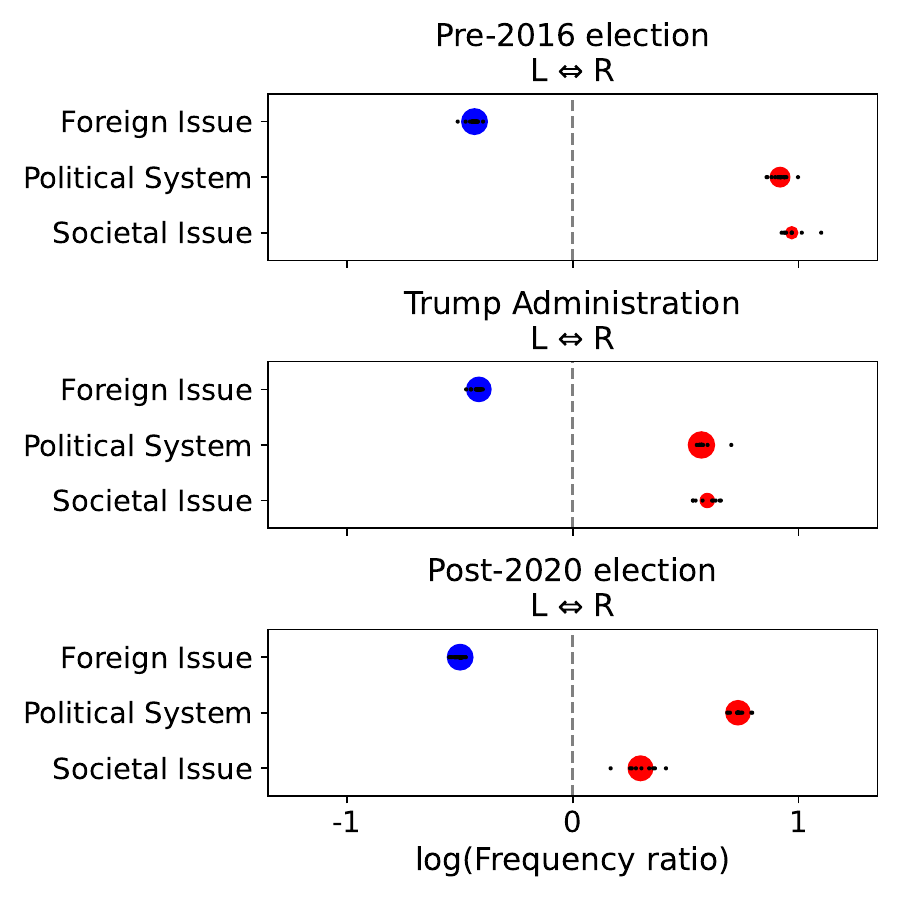}
    \caption{Left versus Right}
    \label{fig:lr_topic}
\end{subfigure}
\hfill
\begin{subfigure}{0.33\textwidth}
    \includegraphics[width=0.85\textwidth]{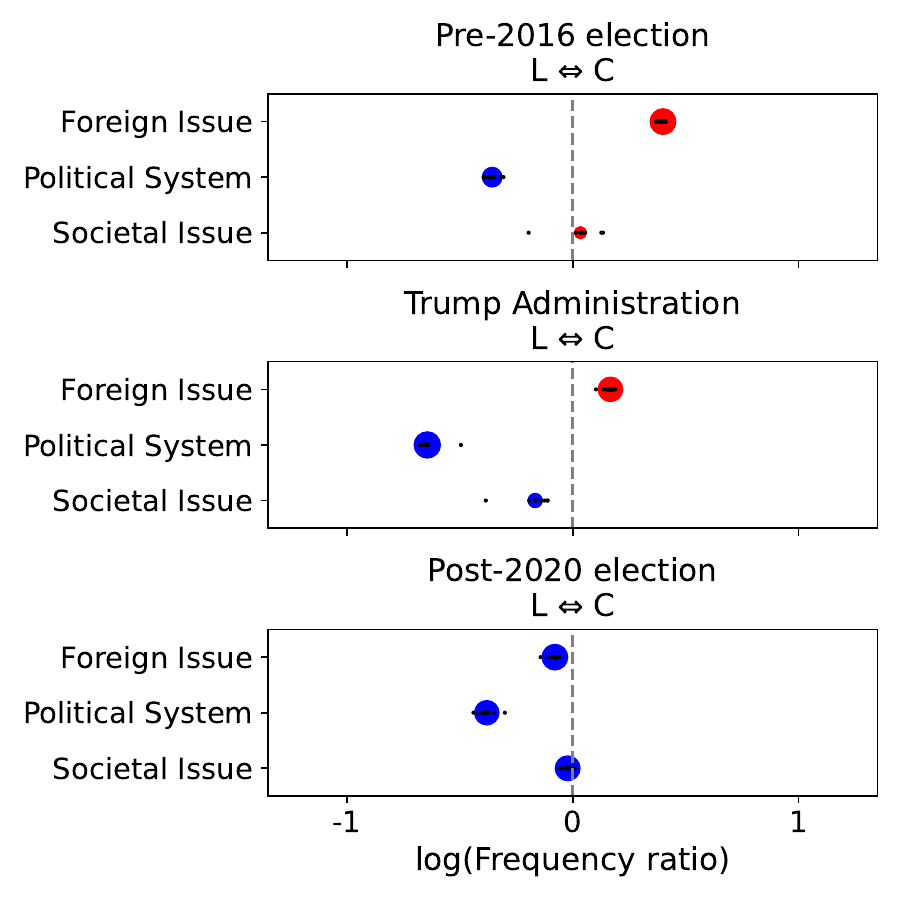}
    \caption{Left versus Central}
    \label{fig:lc_topic}
\end{subfigure}
\hfill
\begin{subfigure}{0.33\textwidth}
    \includegraphics[width=0.85\textwidth]{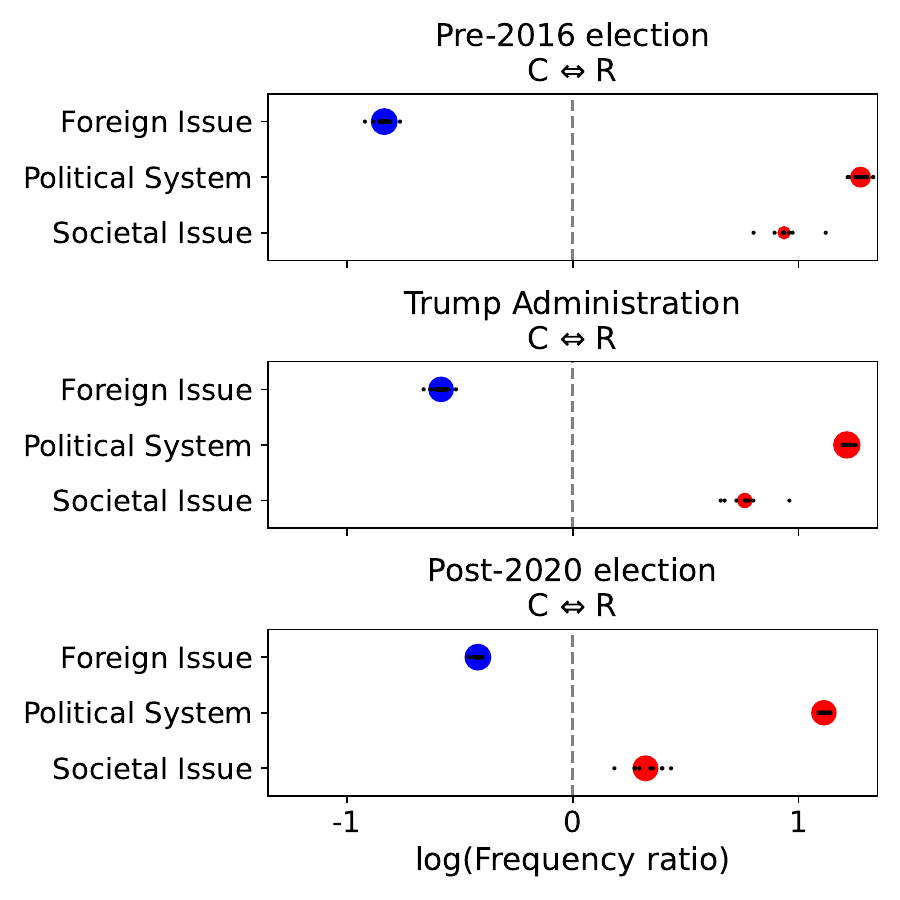}
    \caption{Central versus Right}
    \label{fig:cr_topic}
\end{subfigure}
\caption{Relative frequency for each of three major topics by different media groups. Farther to the right on each plot represents higher frequency by the media on the right-hand side, and vice versa. The circle size represents the overall frequency of the topic. For each topic, we leave one word out and plot the log of frequency ratio again which is then plotted in black.}
\label{fig:topic}
\end{figure*}

Based on the important terms we identify in Table~\ref{tab:impact_word}, we discover three major topics - ``foreign issue'', ``political system'', and ``society issue'', of three media groups across different periods. To better understand the topic divergence among the three media groups, we first measure the difference in topic distributions. For each topic, we curate a list of keywords from the most important terms in Table~\ref{tab:impact_word} (see \textit{Appendix} for details). We then use the keyword list to identify the corresponding topic of each news title. To analyze the distribution divergence among different media groups, the log of frequency ratio is calculated as Eq.~\ref{eq:log} and shown in Figure~\ref{fig:topic}. Following \citet{card2022computational}, we use the log of frequency ratio instead of frequency ratio for a better visualization.
\begin{equation}\label{eq:log}
    \log\frac{F_{LHS}}{F_{RHS}}
\end{equation}
$F_{LHS}$ and $F_{RHS}$ represent the frequency of the media on the left-hand side and right-hand side, respectively. A larger log of frequency ratio means that this topic is mentioned more frequently by the media on the right-hand side and vice versa. The circle size indicates the overall frequency of this topic. Figure~\ref{fig:topic} shows the relative frequency for each of the three topics - ``foreign issue'', ``political system'', and ``societal issue'' among different media groups (Figure~\ref{fig:lr_topic}: {\tt Left} versus {\tt Right}, Figure~\ref{fig:lc_topic}: {\tt Left} versus {\tt Central}, Figure~\ref{fig:cr_topic}: {\tt Central} versus {\tt Right}). To ensure the robustness of our method, we leave one word out of the keyword list and conduct the same analysis again. The new values are plotted as black dots in Figure~\ref{fig:topic}.

In almost all comparisons, we find much difference in the topic distributions. Compared to {\tt Left} and {\tt Central} media, {\tt Right} media pays more attention to the political system and societal issues between 2014 to 2022. {\tt Central} media emphasizes the coverage of foreign issues.

When comparing the difference in topic distributions across the three periods, we observe that the attentions to these three topics of all media groups tend to be more similar where the log of relative frequency is closer to zero.

In terms of the general topic distribution, we find all media groups cover more news about societal issues. More importantly, the difference among the distributions of societal issue topic is reduced in the post-2020 election period.  

\begin{figure}[t]
    \centering
    \includegraphics[width =0.9\linewidth]{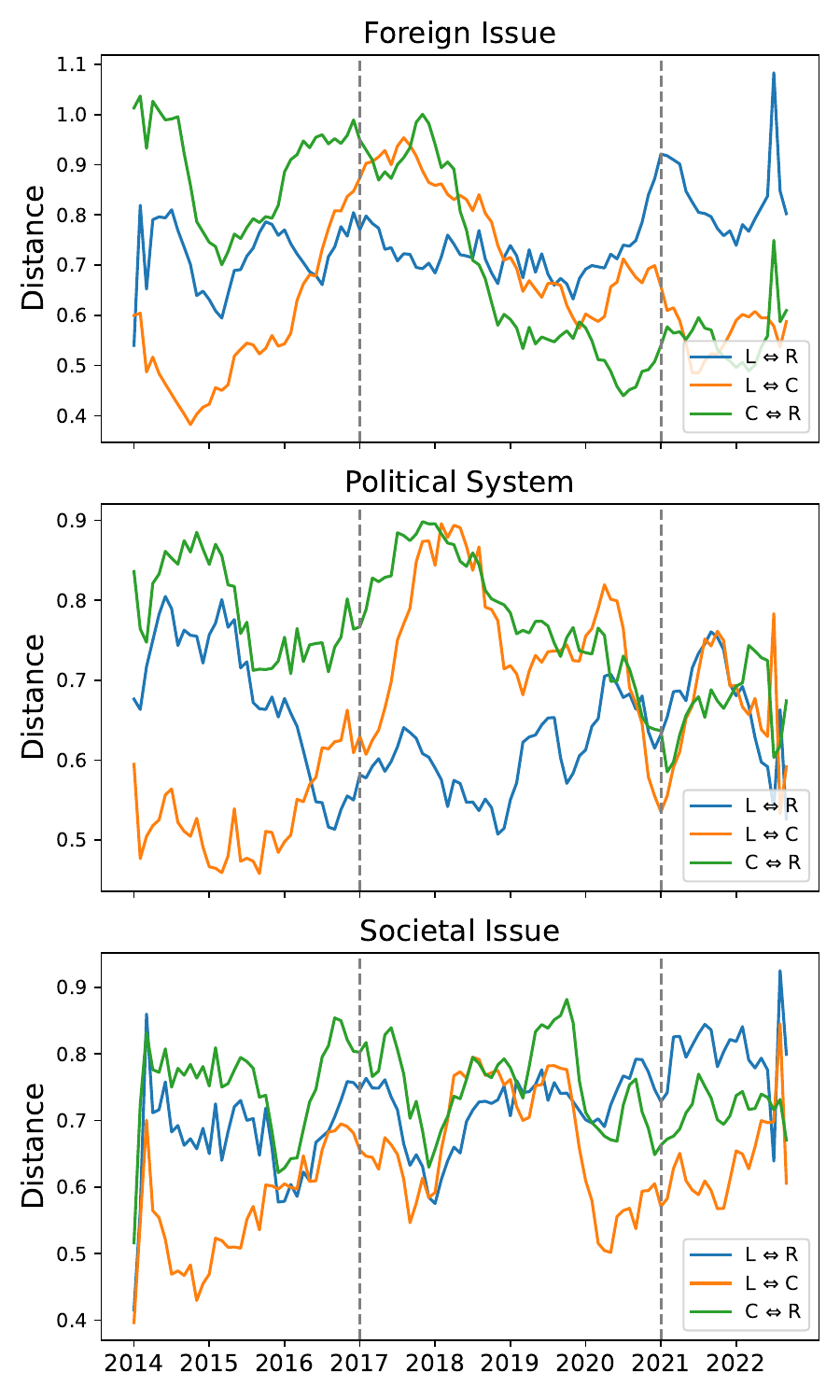}
    \caption{Linguistic difference among different media groups about foreign issues, political systems, and societal issues. L: Left, C: Central, R: Right.}
    \label{fig:linguistic}
\end{figure}

\subsubsection{Linguistic difference}
To estimate how different media groups portray the same topic, we leverage Linguistic Inquiry and Word Count (LIWC)~\cite{pennebaker2001linguistic}. LIWC is a lexicon-based language analysis tool that measures the linguistic features of people's language. In a nutshell, LIWC takes a piece of text data and counts the number of words that fall into each of the pre-defined categories such as emotions, thinking styles, and social concerns that characterizes authors' psychological states based on its dictionary of words and their associated linguistic scores. For instance, LIWC may categorize a word as indicating a positive emotion, such as ``happy''. If the number of ``happy'' is large in that piece of text data, LIWC will output a higher score for positive emotion. In total, LIWC outputs 95 different linguistic features. In our study, we use all of them. Specifically, for each title, we employ LIWC to output a 95-\textit{d} vector of linguistic features.

To ensure the robustness of LIWC on short text such as news titles, we concatenate all the titles of one media group about one topic in a certain period, following \citet{chen2021fine}. Next, because of the scale differences among the linguistic features, we standardize the features. The linguistic difference between any pair of the three media groups is then measured using the cosine distance. The results smoothed using a 7-month moving average are shown in Figure~\ref{fig:linguistic}. Overall, we find three distinct patterns in the linguistic difference between the titles of these topics.

With respect to foreign issues, we find that the linguistic difference between {\tt Right} and {\tt Central} media becomes smaller (green line) while both of them are linguistically more distinct from {\tt Left} media. In contrast, around 2014-2015, there was a smaller difference between {\tt Left} and {\tt Central} than between either of them and {\tt Right}.

We identify two major patterns in the linguistic difference in the coverage of political systems. First, except for the linguistic difference between {\tt Left} and {\tt Central}, the linguistic difference between the other pairs of the three media groups gradually decreases over time.  Second, roughly around the 2016 Presidential Election, the linguistic difference between {\tt Left} and {\tt Right} is almost consistently the lowest compared to the difference between either of them and {\tt Central}, indicating a more similar linguistic style between {\tt Left} and {\tt Right} when they cover news about the political systems (\eg election and politicians).

The linguistic difference in terms of societal issues demonstrates a seasonal pattern where the peak in the linguistic difference occurs approximately \textbf{after} each election (2015, 2017, 2019, 2021). We hypothesize that it is because the societal issues are mainly discussions about policies that are closely related to voters' rights. The different focuses inside the policies are captured by LIWC through the titles. Further, the difference between {\tt Left} and {\tt Central} media is smaller before the 2016 Presidential Election (orange line) compared to the other two cases. However, since then, the differences between the three cases evolve similarly.

\section{Discussions and Conclusions}

Increased partisanship in news can result in enhanced polarization which undermines democracy and has a substantial negative impact across the entire society~\cite{vincent2018crowdsourced,minar2018violence, recuero2020hyperpartisanship, conover2011political,weld2021political, lyu2022understanding}. For instance, researchers found that online partisan media can lead to a lasting and meaningful decrease in trust in the mainstream media~\cite{guess2021consequences}. \citet{recuero2020hyperpartisanship} investigated the role of hyperpartisanship and polarization on Twitter during the 2018 Brazilian presidential Election and found that as the centrality of the hyperpartisan news outlets grows, the mainstream news outlets become more and more biased. 

In this study, we have developed a new dataset for hyperpartisan title detection that allows us to quantify the extent and dynamics of hyperpartisanship in news titles. We consider two types of news titles: (1) the ones that express a one-sided opinion against a policy, apolitical party, or a politician in a biased, inflammatory, and aggressive tone, and (2) the ones that describe a confrontation or a conflict between opposing parties indicating the underlying political climate is polarized. Different from previous studies that only consider the first type of news~\cite{kiesel-etal-2019-semeval, vincent2018crowdsourced,potthast-etal-2018-stylometric}, we include both types as they could both lead to an increase in perceived polarization~\cite{yang2016others,fiorina2005culture,levendusky2016does}.

We find that the {\tt Right} media more often use hyperpartisan titles compared to the {\tt Left} and {\tt Central} media. The proportions of hyperpartisan titles of all media bias groups increased roughly around the 2016 Presidential Election and decreased after the 2020 Presidential Election. A strong association between elections and hyperpartisanship is observed, which is \textit{also reported} by \citet{lin2011more}. Interestingly, the relative change in the proportion of hyperpartisan titles of the {\tt Left} media is the most compared to the {\tt Right} and {\tt Central} media.

Using the logistic regression models with the Shapley values, we find that although the specific words that are suggestive of hyperpartisanship or non-hyperpartisanship in news titles across three media bias groups are \textit{not necessarily the same}, three common topics are identified including \textit{foreign issues}, \textit{political systems}, and \textit{societal issues}. 

Comparisons between any pairs of the {\tt Left}, {\tt Central}, and {\tt Right} media are made in terms of topic frequencies. Societal issues have drawn more attention from all media groups since the start of the Biden Administration. We hypothesize that it is related to the Biden-Harris Administration Immediate Priorities which is intended to address societal issues such as the COVID-19 pandemic, education, climate change, and so on.\footnote{\url{https://www.whitehouse.gov/priorities/}}

Further linguistic analysis reveals three distinct patterns in how different media portrays these topics. First, when it comes to foreign issues, the linguistic difference between {\tt Left} and either {\tt Right} or {\tt Central} media becomes larger while the difference between {\tt Right} and {\tt Central} media is smaller. We hypothesize that the linguistic difference may be suggestive of the conflicting partisan priorities for U.S. foreign policy. For instance, Republicans are more likely to prioritize reducing illegal immigration while Democrats tend to prioritize dealing with global climate change, according to a survey by Pew Research Center.\footnote{\url{https://www.pewresearch.org/politics/2018/11/29/conflicting-partisan-priorities-for-u-s-foreign-policy/}} Second, the linguistic difference in the titles about political systems becomes smaller among all media groups except for the difference between {\tt Left} and {\tt Central}. Third, seasonality is observed in the linguistic difference when depicting societal issues. The difference grows around the start of elections and decreases after the election results are confirmed. We think it is expected since societal issues (\eg education, health) are closely related to voters' rights. It is noteworthy that the linguistic difference between the {\tt Left} and {\tt Central} groups increased for all three topics.

Our computational assessment reveals several new or nuanced aspects about {\it both the extent and dynamics} of hyperpartisanship in news titles. According to the results of the topic analysis, we derive insights into the reasons for the long- and short-term changes. Short-term changes are found related to elections while long-term changes might imply the underlying shifting priorities for the wide range of policies. We use mainstream media news as our study sample. It is worth noting that a similar divergence between different politically leaning entities is also identified in studies that are based on social media data~\cite{waller2021quantifying} or congressional and presidential speeches~\cite{card2022computational}. 

\section{Limitations}
Our study has several limitations. First, there are more and newer news organizations that could be worth investigating. However, given our focus on analyzing the news headlines over time, it is essential that the media we consider should provide publicly available access to their news titles for the entire study period. Other media outlets may not provide such online archives or APIs. It is important to acknowledge that these alternative news sources  often disproportionately feature hyperpartisan titles. Despite this limitation, our experiments suggest that analyzing the behavior of newer outlets would provide a more holistic understanding of media hyperpartisanship and represent a promising avenue for future research. Second, while online news is an important source of news for many people, there are other ways in which individuals consume news. It is crucial to examine media hyperpartisanship in other forms of media as well.

\section*{Acknowledgments}
We are grateful to Ying Zhou, Weihong Qi, Prof. Mayya Komisarchik, and Prof. Anson Kahng for their constructive suggestions. 

\bibliography{aaai24}

\subsection{Paper Checklist}

\begin{enumerate}

\item For most authors...
\begin{enumerate}
    \item  Would answering this research question advance science without violating social contracts, such as violating privacy norms, perpetuating unfair profiling, exacerbating the socio-economic divide, or implying disrespect to societies or cultures?
    \answerYes{Yes.}
  \item Do your main claims in the abstract and introduction accurately reflect the paper's contributions and scope?
    \answerYes{Yes.}
   \item Do you clarify how the proposed methodological approach is appropriate for the claims made? 
    \answerYes{Yes.}
   \item Do you clarify what are possible artifacts in the data used, given population-specific distributions?
    \answerYes{Yes.}
  \item Did you describe the limitations of your work?
    \answerYes{Yes.}
  \item Did you discuss any potential negative societal impacts of your work?
    \answerYes{Yes.}
      \item Did you discuss any potential misuse of your work?
    \answerYes{Yes.}
    \item Did you describe steps taken to prevent or mitigate potential negative outcomes of the research, such as data and model documentation, data anonymization, responsible release, access control, and the reproducibility of findings?
    \answerYes{Yes.}
  \item Have you read the ethics review guidelines and ensured that your paper conforms to them?
    \answerYes{Yes.}
\end{enumerate}

\item Additionally, if your study involves hypotheses testing...
\begin{enumerate}
  \item Did you clearly state the assumptions underlying all theoretical results?
    \answerNA{NA.}
  \item Have you provided justifications for all theoretical results?
    \answerNA{NA.}
  \item Did you discuss competing hypotheses or theories that might challenge or complement your theoretical results?
    \answerNA{NA.}
  \item Have you considered alternative mechanisms or explanations that might account for the same outcomes observed in your study?
    \answerNA{NA.}
  \item Did you address potential biases or limitations in your theoretical framework?
    \answerNA{NA.}
  \item Have you related your theoretical results to the existing literature in social science?
    \answerNA{NA.}
  \item Did you discuss the implications of your theoretical results for policy, practice, or further research in the social science domain?
   \answerNA{NA.}
\end{enumerate}

\item Additionally, if you are including theoretical proofs...
\begin{enumerate}
  \item Did you state the full set of assumptions of all theoretical results?
    \answerNA{NA.}
	\item Did you include complete proofs of all theoretical results?
    \answerNA{NA.}
\end{enumerate}

\item Additionally, if you ran machine learning experiments...
\begin{enumerate}
  \item Did you include the code, data, and instructions needed to reproduce the main experimental results (either in the supplemental material or as a URL)?
    \answerYes{Yes.}
  \item Did you specify all the training details (e.g., data splits, hyperparameters, how they were chosen)?
    \answerYes{Yes.}
     \item Did you report error bars (e.g., with respect to the random seed after running experiments multiple times)?
    \answerNA{NA.}
	\item Did you include the total amount of compute and the type of resources used (e.g., type of GPUs, internal cluster, or cloud provider)?
   \answerNA{NA.}
     \item Do you justify how the proposed evaluation is sufficient and appropriate to the claims made? 
    \answerYes{Yes.}
     \item Do you discuss what is ``the cost`` of misclassification and fault (in)tolerance?
    \answerNA{NA.}
  
\end{enumerate}

\item Additionally, if you are using existing assets (e.g., code, data, models) or curating/releasing new assets, \textbf{without compromising anonymity}...
\begin{enumerate}
  \item If your work uses existing assets, did you cite the creators?
    \answerYes{Yes.}
  \item Did you mention the license of the assets?
    \answerNA{NA.}
  \item Did you include any new assets in the supplemental material or as a URL?
    \answerYes{Yes.}
  \item Did you discuss whether and how consent was obtained from people whose data you're using/curating?
    \answerYes{Yes.}
  \item Did you discuss whether the data you are using/curating contains personally identifiable information or offensive content?
    \answerYes{Yes.}
\item If you are curating or releasing new datasets, did you discuss how you intend to make your datasets FAIR (see \citet{fair})?
\answerYes{Yes.}
\item If you are curating or releasing new datasets, did you create a Datasheet for the Dataset (see \citet{gebru2021datasheets})? 
\answerYes{Yes.}
\end{enumerate}

\item Additionally, if you used crowdsourcing or conducted research with human subjects, \textbf{without compromising anonymity}...
\begin{enumerate}
  \item Did you include the full text of instructions given to participants and screenshots?
    \answerNA{NA.}
  \item Did you describe any potential participant risks, with mentions of Institutional Review Board (IRB) approvals?
    \answerNA{NA.}
  \item Did you include the estimated hourly wage paid to participants and the total amount spent on participant compensation?
    \answerNA{NA.}
   \item Did you discuss how data is stored, shared, and deidentified?
   \answerNA{NA.}
\end{enumerate}

\end{enumerate}

\appendix

\section{Appendix}

\subsection{Additional Dataset Statistics}
Table~\ref{tab:breakdown} shows the breakdown of the titles posted by the nine media outlets. Numbers in parentheses indicate percentages.

\begin{table*}[t]
    \centering
    \begin{tabular}{lccc}
    \toprule[1.1pt]
                          & Hyperpartisan & Non-hyperpartisan & Total   \\
                          \midrule
Bloomberg                 & 10,964 (0.08) & 120,451 (0.92)    & 131,415 \\
CNN                       & 56,519 (0.21) & 213,554 (0.79)    & 270,073 \\
NBC                       & 26,063 (0.20) & 104,585 (0.80)    & 130,648 \\
The New York Times        & 61,492 (0.11) & 492,763 (0.89)    & 554,255 \\
Christian Science Monitor & 11,674 (0.16) & 62,985 (0.84)     & 74,659  \\
Wall Street Journal       & 35,104 (0.08) & 415,174 (0.92)    & 450,278 \\
Reason                    & 17,511 (0.26) & 49,611 (0.74)     & 67,122  \\
The Federalist            & 13,372 (0.39) & 21,080 (0.61)     & 34,452  \\
Washintong Time           & 35,993 (0.33) & 73,729 (0.67)     & 109,722\\
\bottomrule[1.1pt]
\end{tabular}
    \caption{Dataset statistics.}
    \label{tab:breakdown}
\end{table*}

\subsection{{\tt Politics} Subset}
To curate the training set for political title detection, the three annotators who have labeled our hyperpartisan news title dataset further independently label each title as political or not. They are asked to label 1 if the title summarizes political news, and 0 if not. A total of 2,200 titles are labeled. The final label of each title is assigned with the consensus votes from three annotators. Inter-annotator agreement measured by Cohen's Kappa is 0.69, suggesting a substantial agreement~\cite{landis1977measurement}. A different BERT-base model is fine-tuned using this labeled dataset with the same implementation details. On the 200 external validation titles, the accuracy, precision, recall, and $F_1$ scores are 0.91, 0.91, 0.93, and 0.92, respectively. We apply the fine-tuned model to the remaining titles and create the {\tt politics} subset. Table~\ref{tab:data_dist_politics} shows the summary statistics.

\begin{table*}[t]
\centering
\small
\setlength{\tabcolsep}{5pt}
\begin{tabular}{l|l|ccccccccc}
\toprule[1.1pt]
                                  &                                    & 2014 & 2015 & 2016 & 2017 & 2018 & 2019 & 2020 & 2021 & 2022 \\ \midrule
\multirow{2}{*}{\begin{tabular}[c]{@{}l@{}}Manually \\ labeled\end{tabular}} & \# political titles     & 135   & 153     & 162     & 176     & 168     & 167     & 175     & 157     & 135     \\
                                  & \# non-political titles & 85     & 104     & 96     & 86     & 86     & 79     & 81     & 75     & 80     \\ \midrule
\multirow{2}{*}{\begin{tabular}[c]{@{}l@{}}Model \\ labeled\end{tabular}}    & \# political titles     &      61,734 & 62,371 & 72,226 & 65,675 & 62,170 & 61,875 & 64,156 & 57,067 & 52,447     \\
                                  & \# non-political titles &      191,553 & 169,511 & 155,819 & 131,163 & 125,740 & 117,415 & 125,783 & 131,411 & 114,508    \\
                                  \bottomrule[1.1pt]
\end{tabular}
 \caption{Summary statistics of the dataset (political and non-political). The total number of manually labeled titles is 2,200. The total number of titles that are labeled by the model is 1.8 million.}
    \label{tab:data_dist_politics}
\end{table*}

\subsection{Implementation Details}
\subsubsection{Text preprocessing}
We clean our dataset with the following process. First, we remove all Unicode characters and punctuation symbols. We then transform all the text into lowercase. Additionally, we tokenize and lemmatize each word. Finally, the stop words are removed. Tokenization and lemmatization are integrated by using functions from the {\tt nltk} library~\cite{bird2009natural}.

\subsubsection{BERT-base model}
 We fine-tune a vanilla BERT-base model on 2,000 manually labeled titles for 15 epochs and evaluate our model on an external validation set with 200 titles. We choose 32 as the batch size for the training set and 200 as the batch size for the validation set. We select the Adam optimizer~\cite{kingma2014adam} with $\beta_1 = 0.9$, $\beta_2 = 0.999$, and $\epsilon = 10^{-8}$. The learning rate is $2 \times 10^{-5}$. 

\subsubsection{Bag-of-words model}
We first identify all the bigrams in our training set. We then build the co-occurrence matrix. The feature is represented by the mean of word vectors, which are the row vectors from the co-occurrence matrix. After extracting the feature representation of the title, we train the model using an XGBoost classifier with 200 estimators.  

\subsubsection{TF-IDF model}
We exploit the TF-IDF feature extractor from the {\tt sklearn} library~\cite{scikit-learn}. Next, the TF-IDF features are calculated and fed into an XGBoost classifier with 200 estimators.

\subsubsection{Word2Vec model}
We implement the Word2Vec model using the {\tt Gensim} library~\cite{rehurek2011gensim}, which offers open-source natural language processing models. The title representation is the average of each word vector. We train the model using an XGBoost classifier with 200 estimators.

\subsection{Validity Verification for Hyperpartisan Title Detection}
\begin{figure*}[t]
\centering
\begin{subfigure}{0.45\textwidth}
    \includegraphics[width=\textwidth]{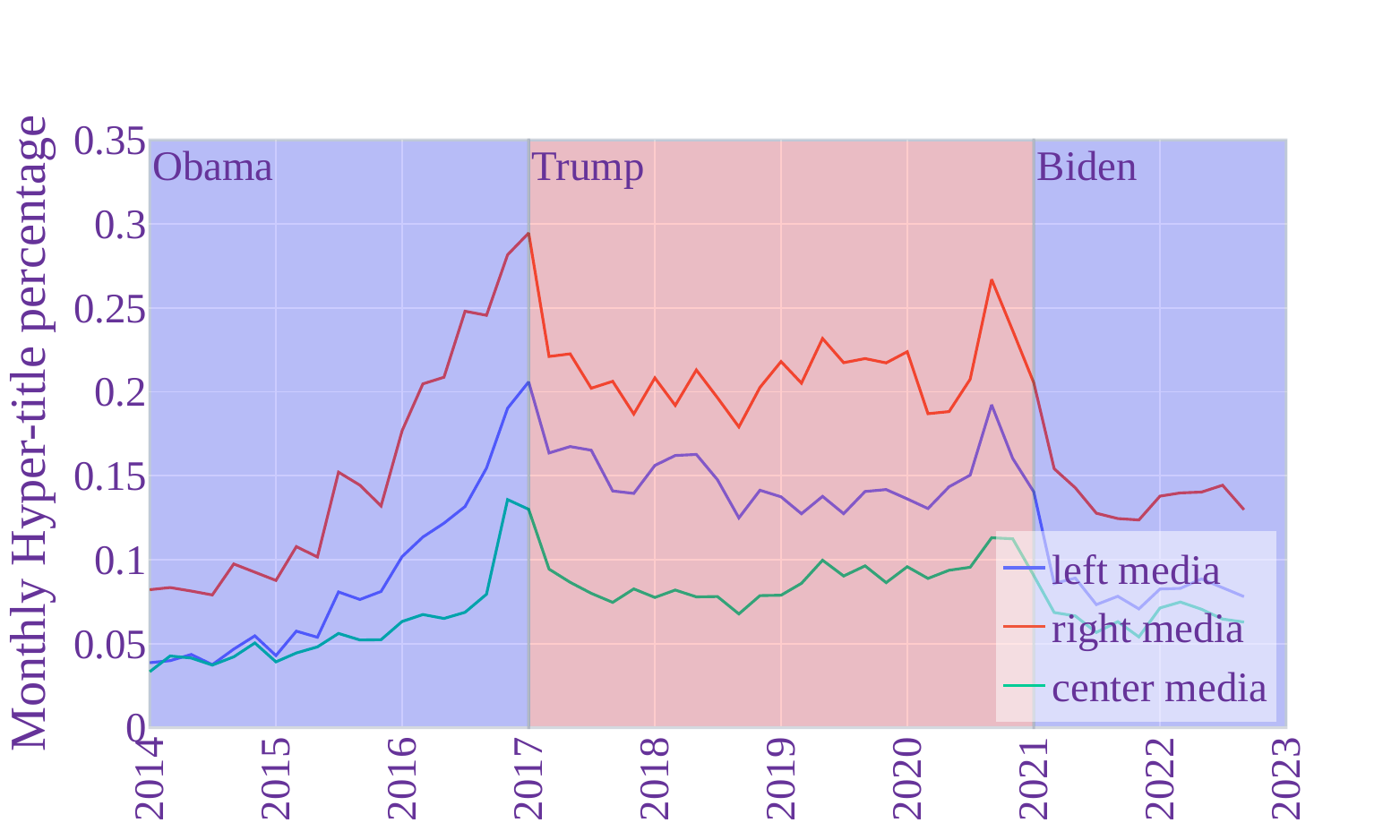}
    \caption{\small Percentage of hyperpartisan titles.}
    \label{fig:first_overall_bow}
\end{subfigure}
\hfill
\begin{subfigure}{0.45\textwidth}
    \includegraphics[width=\textwidth]{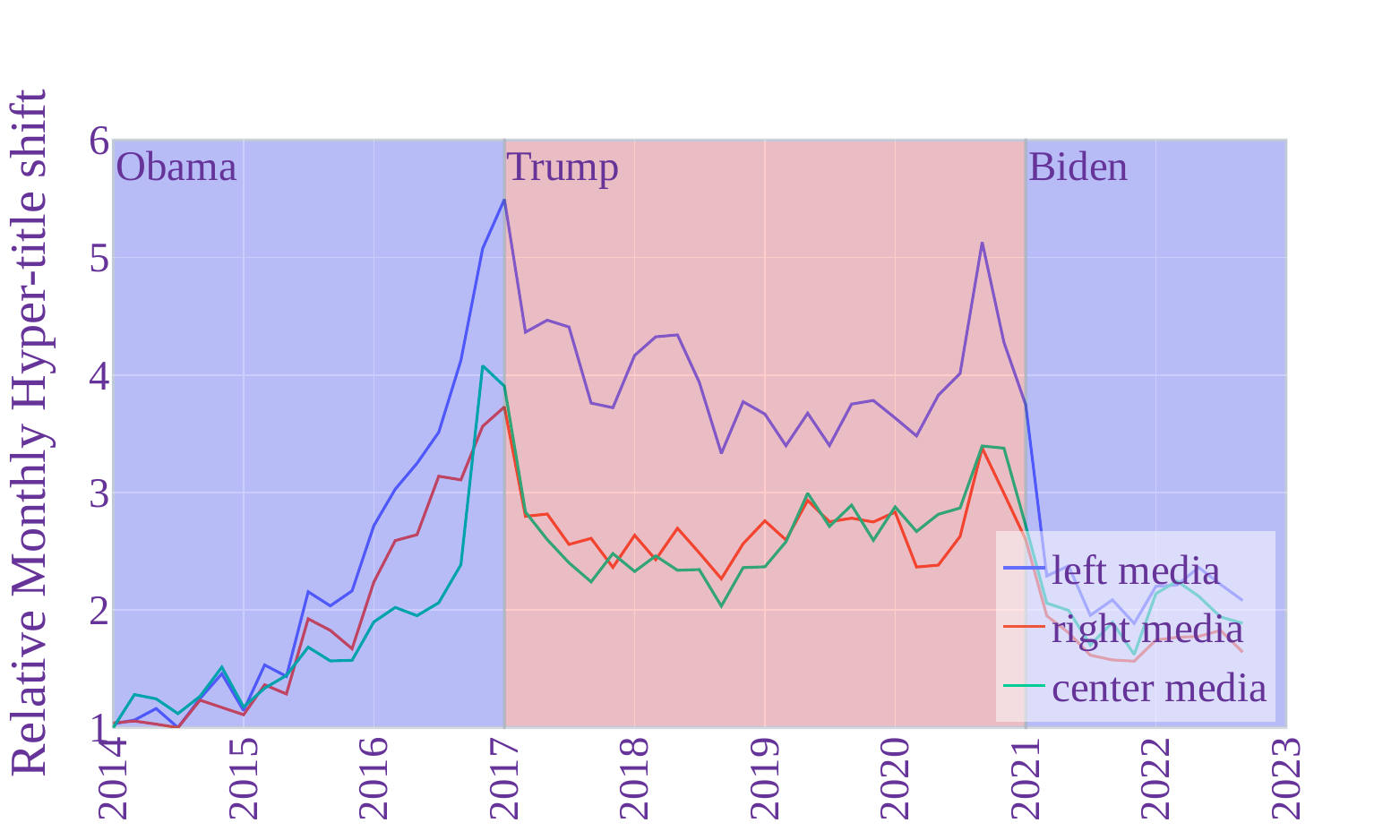}
    \caption{\small Relative change in percentage.}
    \label{fig:second_overall_bow}
\end{subfigure}
\caption{Usage of hyperpartisan titles of Left, Central, and Right media between 2014 to 2022. Replication of Figure~\ref{fig:overall} in the main paper, using XGBoost classifiers with bag-of-words features.}
\label{fig:overall_bow}
\end{figure*}

Pre-trained models may introduce unknown bias from the pre-training data~\cite{card2022computational}. To verify that our results (Figure~\ref{fig:overall} in the main paper) are not significantly influenced by BERT-base, we repeat the hyperpartisanship inference for all the 1.8 million titles using XGBoost classifiers with bag-of-words features and replicate Figure~\ref{fig:overall}. The replication is shown in Figure~\ref{fig:overall_bow}. We find that the overall characteristics and trends do not change. 

\subsection{Search Keywords of Each Topic}
For each of the three topics, we curate a list of related keywords and then use this list to classify each title into its corresponding topic. The noun and adjective of each foreign country\footnote{\url{https://www.englishclub.com/vocabulary/world-countries-nationality.php}} are used as the search keywords to identify the \textit{foreign issue} topic. For \textit{political system} and \textit{societal issue}, one researcher read the words listed in Table~\ref{tab:impact_word} and their corresponding sample titles. The researcher then categorizes them into three groups:(1) political system-related, (2) societal issue-related, and (3) neither. In particular, for each word, twenty titles that contain this word are sampled. If more than 75\% of the samples (\ie titles) belong to either political system-related or societal issue-related, this word is categorized in that group. If neither category is dominant (\ie whose proportion of titles is equal to or greater than 75\%) in these samples, the word will not be included in any keyword lists. As a result, the keyword list used to detect the titles about  \textit{political system} is composed of ``Trump'', ``Obama'', ``Obamacare'', ``Clinton'', ``democrat'', ``republican'', ``GOP'', ``debate'', ``campaign'', ``party'', ``vote'', ``election'', ``Biden'',  ``senate''. The keyword list used to detect the titles regarding \textit{societal issue} includes ``COVID'', ``pandemic'', ``Coronavirus'', ``gun'', ``law'', ``supreme'', ``school'', ``tax'', ``climate''. Note that all words are converted to lowercase during the search so it is not case-sensitive.

To verify the performance of this keyword search method, two annotators independently label 50 sample titles of each topic. The Cohen's Kappa for \textit{foreign issue} and \textit{societal issue} is 0.88 and 0.43, respectively. When the two annotators disagree on a title, we ask them to engage in a thorough discussion to reach a consensus. These labels are then used to verify the keyword search performance. Note that the labels we use to verify the performance of detecting titles belonging to \textit{political system} are from the {\tt politics} subset we construct before. Table~\ref{tab:performance_keyword} shows the performance of the keyword search method. Although the accuracy for \textit{societal issue} detection is not very high, the keyword search method performs well in \textit{foreign issue} and \textit{political system} detection. To further ensure the robustness of our method, we leave one word out of the keyword list and conduct the same analysis on topic distribution again. The new log of frequency ratio is plotted as black dots in Figure~\ref{fig:topic}. We can observe that the patterns mostly remain the same.

\begin{table}[t]
\centering
\setlength{\tabcolsep}{5pt}
\begin{tabular}{ccccc}
\toprule[1.1pt]
Model   & Accuracy & Precision & Recall & $F_1$ \\
\midrule
foreign issue &  0.96       &  0.92        & 1.00       & 0.96   \\
political system & 0.92 & 0.96 & 0.89 & 0.92\\
societal issue   & 0.72         &  0.60         &  0.79      & 0.68  \\
\bottomrule[1.1pt]
\end{tabular}
\caption{Performance of the keyword search method for topic detection.}
\label{tab:performance_keyword}
\end{table}

\subsection{Further Discussion on Potential Broader Impact and Ethical Considerations}
Different from previous studies of hyperpartisanship in news, we introduce a second type of news which describes events that indicate an underlying polarized climate. One potential negative outcome could be that the findings of our study may be interpreted or quoted based on partial statements or without caution. Therefore, we discuss the limitations and how to interpret the results of our study. In terms of the ethical considerations about our dataset, all raw data from the selected mainstream media are publicly available. We develop this dataset to complement the hyperpartisan news datasets of previous studies. 

\end{document}